%% file: paper.tex
\documentclass[conference]{IEEEtran}

\usepackage{cite}
\usepackage{amsmath,amssymb,amsfonts}
\usepackage{algorithmic}
\usepackage{enumerate}
\usepackage{enumitem}
\usepackage{graphicx}
\usepackage{textcomp}
\usepackage{xcolor}

\begin{document}

\title{A Permuted Autoregressive Approach to Word-Level Recognition for Urdu Digital Text}

\input{authors}

\input{sections/0_abstract}

\section{Introduction}
\input{sections/1_introduction}

\section{Related Work}
\input{sections/2_literature}

\section{Data Preprocessing}
\input{sections/3_data}

\section{Experimental Setup}
\input{sections/4_experiment}

\section{Results And Analysis}
\input{sections/5_results}

\section{Conclusion And Future Recommendations}
\input{sections/6_conclusion}

\bibliographystyle{IEEEtran}
\bibliography{references}

\end{document}

%% file: authors.tex
\author{
    \IEEEauthorblockN{Ahmed Mustafa}
    \IEEEauthorblockA{\textit{Department of Robotics \& Intelligent Machines Engineering}\\
    \textit{National University of Sciences \& Technology}\\
    Islamabad, Pakistan\\
    amustafa.rime20smme@student.nust.edu.pk}

    \and
    \IEEEauthorblockN{Muhammad Tahir Rafique}
    \IEEEauthorblockA{\textit{Department of Robotics \& Intelligent Machines Engineering}\\
    \textit{National University of Sciences \& Technology}\\
    Islamabad, Pakistan\\
    mragique.rime17smme@student.nust.edu.pk}

    \and
    \IEEEauthorblockN{Muhammad Ijlal Baig}
    \IEEEauthorblockA{\textit{Department of Electrical Engineering \& Information Technology}\\
    \textit{Technische Universität Darmstadt}\\
    Islamabad, Pakistan\\
    ijlalbaig92@gmail.com}

    \and
    \IEEEauthorblockN{Hasan Sajid}
    \IEEEauthorblockA{\textit{Department of Robotics \& Intelligent Machines Engineering}\\
    \textit{National University of Sciences \& Technology}\\
    Islamabad, Pakistan\\
    hasan.sajid@smme.nust.edu.pk}

    \and
    \IEEEauthorblockN{Muhammad Jawad Khan}
    \IEEEauthorblockA{\textit{Department of Robotics \& Intelligent Machines Engineering}\\
    \textit{National University of Sciences \& Technology}\\
    Islamabad, Pakistan\\
    jawad.khan@smme.nust.edu.pk}

    \and
    \IEEEauthorblockN{Karam Dad Kallu}
    \IEEEauthorblockA{\textit{Department of Robotics \& Intelligent Machines Engineering}\\
    \textit{National University of Sciences \& Technology}\\
    Islamabad, Pakistan\\
    karamdad.kallu@smme.nust.edu.pk}
}

\maketitle

%% file: sections/0_abstract.tex
\begin{abstract}
This research paper introduces a novel word-level Optical Character Recognition (OCR) model specifically designed for digital Urdu text, leveraging transformer-based architectures and attention mechanisms to address the distinct challenges of Urdu script recognition, including its diverse text styles, fonts, and variations. The model employs a permuted autoregressive sequence (PARSeq) architecture, which enhances its performance by enabling context-aware inference and iterative refinement through the training of multiple token permutations. This method allows the model to adeptly manage character reordering and overlapping characters, commonly encountered in Urdu script. Trained on a dataset comprising approximately 160,000 Urdu text images, the model demonstrates a high level of accuracy in capturing the intricacies of Urdu script, achieving a CER of 0.178. Despite ongoing challenges in handling certain text variations, the model exhibits superior accuracy and effectiveness in practical applications. Future work will focus on refining the model through advanced data augmentation techniques and the integration of context-aware language models to further enhance its performance and robustness in Urdu text recognition.
\end{abstract}

\hfill

\begin{IEEEkeywords}
Urdu text recognition, PARSeq, Permutation language modeling, Transformers
\end{IEEEkeywords}

%% file: sections/1_introduction.tex
Optical Character Recognition (OCR) plays a vital role in various applications, such as document analysis, image understanding, and intelligent character recognition. Over the years, significant progress has been made in the field of OCR, with numerous techniques and models developed to tackle the challenges posed by different languages and scripts. However, much of the existing research focuses on widely studied languages like English, Chinese, and French. Urdu, an important language spoken by millions of people, has received limited attention in the context of OCR. With its unique script and inherent complexities, Urdu presents significant challenges for accurate text recognition. This paper aims to bridge this research gap by focusing on Urdu text recognition.

Despite the significant progress in English OCR, the scope of research in this domain is limited when it comes to Urdu, a complex language with unique script characteristics. Recognizing digital Urdu text poses several challenges due to the nature of its connective writing system and the diversity of font styles. The scarcity of research in Urdu OCR demands novel solutions and tailored deep learning architectures that can effectively tackle the complexities of this language. Our research aims to address this gap by designing and developing an optimized deep learning model for Urdu OCR, specifically utilizing the PARSeq architecture \cite{b1} to effectively recognize digital Urdu text. In addition to addressing the core challenges of recognizing digital Urdu text, this research explores the necessary modifications in preprocessing, training, and inference processes to accommodate the nuances of the Urdu script and its linguistic characteristics.

By expanding the scope of research in digital Urdu text recognition, this paper aims to contribute to the development of reliable and robust OCR systems for Urdu. Urdu is one of the most widely spoken languages in South Asia, with a rich literary tradition and cultural heritage. Accurate Urdu text recognition can have profound implications across various domains, including information retrieval, document analysis, and automated tasks such as translation, text-to-speech conversion, and sentiment analysis. Additionally, Urdu OCR has practical applications in sectors like banking, government services, and education, where digitizing and extracting information from Urdu documents can enhance administrative processes, improve communication, and provide better access to services for Urdu-speaking populations. Ultimately, the goal is to unlock the potential of Urdu OCR and integrate it effectively into various digital platforms and real-world applications.

In the subsequent chapters, we will explore the existing literature, present the approaches we employed, discuss the results, and provide insights into future directions for further improvement in Urdu text recognition. Through our research, we aim to make valuable contributions to the field of OCR and facilitate advancements in Urdu language processing.

%% file: sections/2_literature.tex
Optical Character Recognition (OCR) has significantly advanced from traditional computer vision techniques to sophisticated deep learning architectures. Early OCR systems primarily relied on pattern recognition and template matching \cite{b2}, which were limited in handling complex fonts and handwriting variations. The introduction of machine learning techniques, such as Support Vector Machines (SVM), Hidden Markov Models (HMM), and neural networks \cite{b3}, improved OCR by allowing better handling of text variations and increasing adaptability to different types of input. However, these methods often fell short in capturing long-range dependencies in sequential data, limiting their effectiveness in more complex text recognition tasks.

Recurrent Neural Networks (RNNs), particularly Long Short-Term Memory (LSTM) networks \cite{b4}, addressed some of these challenges by preserving contextual information, which is crucial for multi-line and variable-length text recognition. Despite their promise, RNN-based OCR systems struggled with slow inference times due to the sequential nature of character generation. The integration of attention mechanisms \cite{b5} marked a significant advancement, enabling models to focus selectively on relevant image regions, thereby improving recognition accuracy and robustness across varied text conditions, including differences in font size, style, and orientation.

More recently, Non-Autoregressive (NAR) models \cite{b6} have emerged, offering parallel processing capabilities that reduce inference time and enhance efficiency. These models are particularly suitable for real-time applications and large-scale tasks. However, even with these advancements, handling context and global dependencies in long documents remained a challenge, prompting the integration of Transformer networks into OCR research. Transformers, introduced by Vaswani et al. \cite{b7}, demonstrated exceptional capabilities in managing long-range dependencies and capturing global context in sequential data. Transformer-based OCR models have achieved state-of-the-art results, surpassing previous methods in accuracy, robustness, and speed.

While most OCR research has focused on widely-used languages like English, there is increasing recognition of the need for OCR systems tailored to complex scripts such as Urdu. Urdu OCR presents unique challenges due to its cursive nature and diverse ligature forms. Various approaches have been explored, from segmentation-based methods that break down text into individual characters \cite{b8}\cite{b9} to segmentation-free techniques that recognize text at the ligature or word level \cite{b10}.

Segmentation-based approaches, such as those using morphological processing and contour analysis \cite{b8}, simplify the recognition task but face significant challenges in accurately segmenting characters. Techniques like horizontal and vertical projection profiles \cite{b9} have also been employed to address these challenges, achieving varying levels of success. Conversely, segmentation-free approaches handle a larger number of classes due to the complexity of the Urdu script but can be easier to implement as they avoid explicit segmentation. These methods have utilized template matching and stroke-based features for character or ligature recognition \cite{b10}.

Deep learning models have also been leveraged to improve Urdu handwriting recognition, addressing data scarcity issues and effectively handling the shape variations of Urdu characters depending on their context. These models provide robust solutions by learning intricate patterns within the script. Recent studies have introduced convolutional neural networks for feature extraction from handwritten text images, combined with Gated Recurrent Units (GRU) and attention mechanisms for sequence prediction. High-resolution images and BiLSTM layers have also been used to enhance the recognition of printed Urdu text, further emphasizing the importance of high-quality data and advanced architectures in capturing the intricacies of Urdu script and improving overall model performance \cite{b11}.

As previously noted, Transformer networks have made significant advancements in sequence-to-sequence tasks such as OCR. Capturing long-range dependencies and contextual information is essential in OCR, particularly when dealing with complex scripts like Urdu. Transformer-based models offer substantial benefits in this context since the attention mechanisms intrinsic to these architectures allow the model to selectively focus on different segments of the input text, dynamically capturing relevant features. This capability is especially beneficial for recognizing cursive scripts such as Urdu, which involve intricate shape variations and significant contextual dependencies between characters and ligatures. By effectively addressing these challenges, Transformer-based OCR models achieve high levels of accuracy and robustness, demonstrating their efficacy in handling the complexities associated with such scripts \cite{b12}.

In this paper, we delve deeper into the methodologies and experiments involved in developing a robust Urdu text recognition model by exploring the adaptation of language semantics and incorporating the innovations of the PARSeq model \cite{b1}. This research contributes to advancing the state-of-the-art in Urdu text recognition and broadening the scope of OCR applications, particularly for languages with complex scripts.

%% file: sections/3_data.tex
\subsection{Data Collection}
The dataset collection process involved capturing images of digitally printed Urdu text from various sources, such as pages, banners, identity cards, and other printed materials, to ensure a comprehensive representation of the scenarios encountered in Urdu text recognition applications. To enhance the robustness and generalization capabilities of the trained models, various factors, including lighting conditions, camera settings, image resolutions, and zoom levels, were systematically varied during image acquisition. This deliberate variation was intended to improve the model’s ability to generalize across different conditions. The final dataset comprised approximately $160,000$ images of cropped digital Urdu words. For training the OCR model, the dataset was meticulously annotated at the word level to provide precise ground truth labels. Word-level annotation allowed the model to learn the contextual relationships and dependencies among characters within each word, thus enhancing recognition accuracy. Due to the specific nature of this research and confidentiality requirements, the dataset is not publicly available.

\subsection{Image Preprocessing}
Image preprocessing is a crucial step in any vision based machine learning task, and it plays a significant role in Optical Character Recognition (OCR) as well. After data annotation, the dataset underwent several preprocessing steps to enhance the quality and readability of the Urdu text images.
\begin{itemize}
\item Images were processed to reduce noise caused by factors such as sensor noise, compression artifacts, or environmental influences. Techniques like median filtering and Gaussian filtering were applied to suppress noise while preserving text details.
\item Skew refers to the slant or tilt in the orientation of the text, which can adversely affect recognition accuracy. To address this, skew correction algorithms were implemented to align text lines horizontally, thereby improving legibility and reducing recognition errors associated with skewed text.
\item Contrast enhancement techniques were utilized to improve the visibility and contrast of text against varying backgrounds and lighting conditions. These methods adjusted the pixel intensities of images to enhance text visibility, making it more distinguishable from surrounding elements.
\item Data augmentation techniques were employed to increase the diversity and size of the dataset. This process included applying random rotations, translations, scaling, filters, and cropping to the images. Augmentation helped the model generalize better by exposing it to various variations and distortions that may occur in real-world scenarios.
\end{itemize}

By employing these preprocessing techniques, the dataset was thoroughly prepared for training the Urdu text recognition model. The processed dataset provided a robust foundation for the subsequent stages of model development and evaluation, ensuring that the model was trained on high-quality data. This careful preparation was essential for enhancing the model's ability to accurately recognize and interpret Urdu text under various conditions.

\subsection{Label Conversion}
When developing an OCR system for Urdu text, a distinct preprocessing approach is necessary due to the unique characteristics of the Urdu script. Unlike English, where each character retains a consistent form regardless of its position within a word, Urdu characters change shape depending on their placement—initial, medial, final, or isolated. This positional variation adds complexity to the recognition process, as each character can have multiple forms depending on its context. To manage these variations effectively, each distinct shape that a character can assume based on its position within a word is treated as a separate output label during the training process. This differs from simpler scripts, like Latin, where each character typically corresponds to a single label. By assigning separate labels for each positional variant of a character, the number of labels in the model increases significantly beyond the actual number of characters, symbols, and numerals in the Urdu language. This approach allows the model to learn the contextual dependencies and positional characteristics of characters within words, which is crucial for accurate recognition of Urdu text.

%% file: sections/4_experiment.tex
\subsection{Architecture}
At the core of the architecture is the innovative Permutation Autoregressive Sequence (PARSeq) model, which introduces a novel approach to the complexities of Urdu text recognition. While the detailed mechanics of PARSeq are extensively covered in its dedicated research paper \cite{b1}, a brief overview of its adaptation for Urdu OCR is necessary. PARSeq departs from the traditional autoregressive paradigm by using permutation-based language modeling, allowing the model to go beyond the constraints of conventional left-to-right (LTR) sequential predictions. This is particularly advantageous for languages like Urdu, which feature complex ligatures, non-standard character arrangements, and irregular text orientations.

PARSeq’s strength lies in its ability to predict multiple permutations of token orderings, capturing the diverse ways characters can be sequenced. In Urdu OCR, PARSeq modifies how characters, words, and sentences are predicted by accommodating various token permutations, thus inherently adapting to the intricacies of Urdu script and mitigating the limitations of sequential autoregressive models. This flexibility aligns well with the complex nature of Urdu script, enabling the model to consider a wider contextual range when generating predictions.

The implementation of PARSeq involves training the model to predict multiple token permutations. Instead of generating a single sequential output, the model learns various token permutations, capturing different character orders. During inference, this adaptability allows the model to select the most contextually appropriate permutation, aligning with the actual word order. This approach effectively handles character reordering and overlapping characters, key features of Urdu script. For computational efficiency, PARSeq does not utilize all possible permutations during training; instead, it randomly samples a set of $\mathbf{K}$ permutations per training example. This approach optimizes computational resources while ensuring the model can grasp the complexities of Urdu script.

\subsection{Loss Function}
Cross-entropy loss is a widely used objective function in neural network training, especially for classification tasks such as Optical Character Recognition (OCR) in Transformer-based models. In the context of OCR, cross-entropy loss measures the divergence between the predicted probability distribution and the actual distribution of target labels. By penalizing incorrect predictions based on the predicted probabilities, this loss function facilitates the optimization of model parameters to enhance recognition accuracy.

\[\mathbf{L_{CE}} = -\sum_{i=1}^{N} y_i \log(\hat{y}_i)\]

The incorporation of $\mathbf{K}$ permutations in PARSeq model necessitated a modification to the conventional cross-entropy loss function. Traditional autoregressive models rely on a single fixed permutation for likelihood factorization. In contrast, PARSeq involves sampling $\mathbf{K}$ permutations from the set of all $\mathbf{T!}$ possible factorizations, thereby requiring a novel loss formulation that aggregates cross-entropy losses across multiple permutation-derived attention masks:

\[\mathbf{L} = \frac{1}{K}\sum_{k=1}^{K}L_{CE}(y_k,\hat{y})\]

This adjustment allows the model to better capture the complexities of the target script.

\subsection{Model Evaluation}
The evaluation of the proposed architecture's performance centers on its accuracy and effectiveness in recognizing Urdu text. The primary metric used is the Character Error Rate (CER), a standard measure in OCR evaluation. CER quantifies the discrepancy between the predicted text output and the ground truth text by calculating the number of character-level substitutions, deletions, and insertions needed to transform the predicted text into the ground truth. To compute CER, the predicted text is compared with the ground truth at the word level. The CER is then calculated by dividing the total number of character-level errors by the total number of characters in the ground truth text, typically expressed as a percentage. This metric provides a comprehensive assessment of the model’s performance by accounting for character-level errors, thereby allowing for minor deviations in the predicted text as long as the overall accuracy of word recognition is maintained.

\subsection{Training}
The hyperparameters used for training the final model are detailed in Table \ref{table:hyperparameters}.

\input{tables/hyperparameters}

%% file: tables/hyperparameters.tex
\begin{table}[htbp]
\caption{Hyperparameters}
\begin{center}
\begin{tabular}{|c|c|}
\hline
\textbf{Hyperparameter} & \textbf{Value} \\
\hline
Learning rate & $0.1$ \\
\hline
Embedding size & $256$ \\
\hline
Heads & $4$ \\
\hline
Encoder layers & $4$ \\
\hline
Decoder layers & $2$ \\
\hline
Feedforward dimensions & $1024$ \\
\hline
Dropout & $0.3$ \\
\hline
Permutations & $3$ \\
\hline
Batch size & $32$ \\
\hline
Epochs & $7600$ \\
\hline
\end{tabular}
\label{table:hyperparameters}
\end{center}
\end{table}

%% file: sections/5_results.tex
\subsection{Performance}
The proposed architecture was evaluated using a test set of $1470$ digital Urdu text images, achieving an average CER of $0.178$. During training, the model reached a final training loss of $0.0003$ and a validation loss of $0.7$, indicating effective convergence and good generalization to unseen data.

The low CER observed in testing suggests that the model is capable of accurately recognizing and extracting text from digital Urdu images. Notably, the model performed well even under challenging conditions, such as overlapping characters and variable illumination. These results highlight the model’s robustness in handling such difficult cases, showcasing its practical utility for digital text recognition tasks.

Overall, these results demonstrate the model's strong adaptability and proficiency in recognizing and extracting Urdu text, indicating its potential for deployment in real-world applications where text recognition is required.

\subsection{Challenges}
While the model demonstrated strong overall performance, there were some instances where recognition accuracy could be enhanced. These cases of reduced accuracy were primarily due to the following factors:

\begin{enumerate}
\item \textbf{Blurred Images:} Intense blurring can sometimes cause incomplete or distorted character representations, affecting feature extraction and leading to recognition errors.

\item \textbf{Non-Horizontal Orientation:} Text that is not horizontally aligned or appears at various angles can complicate the OCR process, as traditional models are optimized for straight text.

\item \textbf{Background Interference:} Elements such as patterns, lines, or overlapping text can create noise that interferes with character recognition.

\item \textbf{Punctuations:} Punctuation marks that are close to text can sometimes be misinterpreted as part of the characters, leading to inaccuracies.
\end{enumerate}

Figure \ref{fig:challenges} illustrates examples of images that presented these challenges for the model, highlighting specific areas where performance was affected. While these factors did impact accuracy in certain instances, they primarily suggest opportunities for refinement rather than indicating fundamental deficiencies of the model since the observed inaccuracies are likely due to the limited diversity of these specific cases in the training data. By expanding the dataset to include a broader range of such challenging scenarios, it is anticipated that these limitations can be substantially mitigated, thereby enhancing the model's robustness and accuracy in future iterations.

\subsection{Comparative Analysis}
In the context of evaluating our Urdu OCR model's strengths relative to contemporary approaches, several challenges arise due to the absence of standardized benchmarks and limited accessibility to proposed architectures. Research in this domain is predominantly conducted at the university level, where promising models are often confined to academic circles and are not made publicly available. Furthermore, the lack of a consistent evaluation metric complicates direct comparisons between models, making meaningful assessments challenging.  Given these limitations, we opted to evaluate our model's performance against Google Vision OCR, which serves as a widely recognized industry benchmark.

Upon evaluation on the same test dataset, Google Vision OCR recorded a CER of $0.891$, which is notably higher than the $0.178$ CER achieved by our model. The performance gap provides valuable insights into the unique strengths of our model and its effectiveness in managing the specific challenges associated with Urdu OCR. By surpassing a leading commercial OCR system, our model demonstrates its capability in effectively handling Urdu text recognition, positioning it as a strong contender in this field.

%% file: sections/6_conclusion.tex
This paper addresses the challenge of text recognition for digital Urdu text, a domain that has received limited research attention. The primary objective was to develop an OCR model specifically designed for the unique characteristics of the Urdu script, utilizing the PARSeq architecture to effectively manage the complexities associated with Urdu text recognition.

The developed model exhibits notable strengths, particularly in its ability to handle the distinctive challenges posed by Urdu text, such as overlapping characters and variations in font styles. It demonstrates robust performance across a diverse range of text samples, attributed to training on a comprehensive dataset of Urdu text, which enhances the model's accuracy and generalization.

Despite these strengths, the model has certain limitations which could be further optimized by incorporating a wider variety of training examples. Future work can focus on refining the model's performance through advanced data augmentation techniques as well as enhancing its ability to capture contextual relationships between characters in longer text sequences by integrating context-aware language models.

By addressing these limitations, OCR technology for Urdu has the potential to achieve greater accuracy and versatility, thereby contributing to the advancement of multilingual OCR and fostering further innovations in text recognition technology.

\input{figures/challenges}

%% file: figures/challenges.tex
\begin{figure}
    \centering
    \begin{enumerate}[label=(\alph*)]
        \item \raisebox{-37pt}{
            \includegraphics[width=0.95\linewidth, height=50pt]{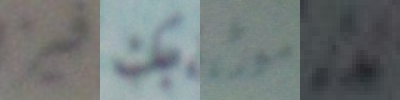}}
            \vspace{2pt}
        \item \raisebox{-37pt}{
            \includegraphics[width=0.95\linewidth, height=50pt]{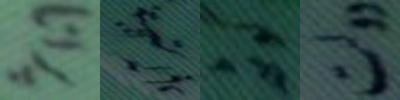}}
            \vspace{2pt}
        \item \raisebox{-37pt}{
            \includegraphics[width=0.95\linewidth, height=50pt]{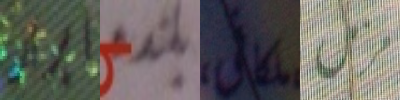}}
            \vspace{2pt}
        \item \raisebox{-37pt}{
            \includegraphics[width=0.95\linewidth, height=50pt]{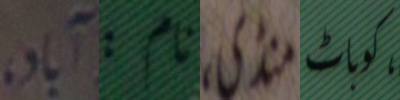}}
    \end{enumerate}
    \caption{
        (a) Blurred Images,
        (b) Non-Horizontal Orientation,
        (c) Background Interference,
        (d) Punctuations
    }
    \label{fig:challenges}
\end{figure}